\documentclass[conference]{IEEEtran}
\IEEEoverridecommandlockouts

\usepackage{cite}
\usepackage{amsmath,amssymb,amsfonts}
\usepackage{algorithmic}
\usepackage{graphicx}
\usepackage{subcaption}
\usepackage{graphicx}
\usepackage{textcomp}
\usepackage{xcolor}
\usepackage{float}
\usepackage{acronym}
\usepackage{siunitx}
\usepackage{booktabs} 
\usepackage{multirow}

\def\BibTeX{{\rm B\kern-.05em{\sc i\kern-.025em b}\kern-.08em
    T\kern-.1667em\lower.7ex\hbox{E}\kern-.125emX}}
\begin{document}
\acrodef{SUMO}{Simulation of Urban Mobility}
\acrodef{C-ITS}{Cooperative Intelligent Transport Systems}
\acrodef{TMS}{Traffic Management System}
\acrodef{ETSI}{European Telecommunications Standards Institute}
\acrodef{RSU}{Road-Side Unit}
\acrodef{MC}{Mobile Cloud}
\acrodef{SPaT}{Signal, Phase and Timing}
\acrodef{SPATEM}{Signal, Phase and Timing Extended Message}
\acrodef{MAPEM}{Map Extended Message}
\acrodef{MAP}{Map Data}
\acrodef{DENM}{Decentralized Environmental Notification Message}
\acrodef{CAM}{Cooperative Awareness Message}
\acrodef{IVIM}{Infrastructure to Vehicle Information Message}
\acrodef{SREM}{Signal Request Extended Message}
\acrodef{SSEM}{Signal Status Extended Message}
\acrodef{CCAM}{Connected, Cooperative and Automated Mobility}
\acrodef{CPM}{Collective Perception Message}
\acrodef{CPS}{Collective Perception Service}
\acrodef{MRM}{Maneuver Recommendation Message}
\acrodef{MCM}{Maneuver Coordination Message}
\acrodef{GLOSA}{Green-Light Optimized Speed Advisory}
\acrodef{PVD}{Probe Vehicle Data}
\acrodef{MQTT}{Message Queuing Telemetry Transport}
\acrodef{NATS}{NATS Messaging}
\acrodef{gRPC}{gRPC Remote Procedure Calls}
\acrodef{PoE}{Power-over-Ethernet}
\acrodef{PPS}{Pulse-per-Second}
\acrodef{REMAS}{Resource Management System for Highly Automated Urban Traffic}
\acrodef{C4CART}{Communications for Connected and Automated Road Traffic}
\acrodef{V2N}{Vehicle-to-Network}
\acrodef{V2V}{Vehicle-to-Vehicle}
\acrodef{V2X}{Vehicle-to-Everything}
\acrodef{V2I}{Vehicle-to-Infrastructure}
\acrodef{C-V2X}{Cellular V2X}
\acrodef{RUMBA}{Road-Side Unit Module Based Architecture}
\acrodef{DOM}{Dynamic Object Map}
\acrodef{DAB}{Digital Audio Broadcasting}
\acrodef{CAV}{Connected Automated Vehicle}
\acrodef{CAVs}{Connected Automated Vehicles}
\acrodef{LIN}{Local-Interconnect Network}
\acrodef{C-BOX}{Connectivity Box}
\acrodef{ITS}{Intelligent Transport System}
\acrodef{ACV}{Automated Connected Vehicle}
\acrodef{NACV}{Non-Automated Connected Vehicle}
\acrodef{HMI}{Human-Machine-Interface}
\acrodef{LTE}{Long Term Evolution}
\acrodef{ADS}{Automated Driving System}
\acrodef{ODD}{Operational Design Domain}
\acrodef{D2D}{Device-to-Device}
\acrodef{BTP}{Basic Transport Protocol}
\acrodef{pHUD}{portable Head-Up Display}
\acrodef{OSM}{OpenStreetMap}
\acrodef{TraCI}{Traffic Control Interface}
\acrodef{LCM}{Lightweight Communications and Marshalling}
\acrodef{OBU}{On-Board Unit}
\acrodef{QoS}{Quality of Service}
\acrodef{VAMOS}{Verkehrs-Analyse-, -Management- und -Optimierungs-System/traffic analysis, management and optimization system}
\acrodef{VRU}{Vulnerable Road User}
\acrodef{VRUs}{Vulnerable Road Users}
\acrodef{LoS}{Level of Service}
\acrodef{HiL}{Hardware-in-the-Loop}
\acrodef{SiL}{Software-in-the-Loop}
\acrodef{TSP}{Traffic Signal Priority request}
\acrodef{EVA}{Emergency Vehicle Approaching}
\acrodef{OTA}{Over-the-Air}
\acrodef{MTU}{Maximum Transfer Unit}
\acrodef{UPER}{Unaligned Packed Encoding Rules}
\acrodef{PRR}{Packet Reception Ratio}
\acrodef{LOS}{Line-of-Sight}
\acrodef{NLOS}{Non-Line-of-Sight}
\acrodef{UWB}{Ultra-Wideband}
\acrodef{C-ACC}{Connected Adaptive Cruise Control}
\acrodef{VuT}{Vehicle under Test}
\acrodef{MPC}{Model predictive control}
\acrodef{CEP}{Circular Error Probable}
\acrodef{BSM}{Basic Safety Message}
\acrodef{ACC}{Adaptive Cruise Control}
\acrodef{CV}{Connected Vehicle}
\acrodef{CAN}{Controller Area Network}
\acrodef{IoT}{Internet of Things}
\acrodef{ABS}{Anti-lock Braking System}
\acrodef{UAV}{Unmanned Aerial Vehicles}
\acrodef{RAN}{Radio Access Network}
\acrodef{KPI}{Key Performance Indicator}
\acrodef{SA}{standalone}
\acrodef{CBR}{Constant Bitrate}
\acrodef{FPS}{Frames Per Second}
\acrodef{RTSP}{Real Time Streaming Protocol}
\acrodef{H.264}{Advanced Video Coding}
\acrodef{CRC}{Cyclic Redundancy Check}
\acrodef{MITM}{Man-in-the-Middle}
\acrodef{SSL}{Secure Socket Layer}
\acrodef{ACL}{Access Control List}
\acrodef{API}{Application Programmable Interface}
\acrodef{VPN}{Virtual Private Network}
\acrodef{L4S}{Low-Latency, Low-Loss, Scalable Throughput}
\acrodef{LSTM}{Long-Short-Term Memory}
\acrodef{ADE}{Average Displacement Error}
\acrodef{FDE}{Final Displacement Error}
\acrodef{VAM}{VRU Awareness Message}
\acrodef{CP}{Collaborative Perception}
\acrodef{GNSS}{Global Navigation Satellite System}
\acrodef{DSRC}{Dedicated Short-Range Communication}
\acrodef{C-V2X}{Cellular-V2X}
\acrodef{NTP}{Network Time Protocol}
\acrodef{PTP}{Precision Time Protocol}
\acrodef{AP}{Average Precision}
\acrodef{IoU}{Intersection over Union}
\acrodef{BEV}{Bird's-Eye View}

\title{VALISENS: A Validated Innovative Multi-Sensor System for Cooperative Automated Driving
\thanks{This research is financially supported by the German Federal Ministry for Economic Affairs and Climate Action (BMWK) under grant number FKZ 19A22009F (VALISENS).}
}

\author{
    \IEEEauthorblockN{Lei Wan\IEEEauthorrefmark{1}\IEEEauthorrefmark{2}, Hannan Ejaz Keen\IEEEauthorrefmark{1}, Prabesh Gupta\IEEEauthorrefmark{3}, Andreas Eich\IEEEauthorrefmark{3}, Michael Klöppel-Gersdorf\IEEEauthorrefmark{4},\\
    Marcel Kettelgerdes\IEEEauthorrefmark{4}, Maximilian Bialdyga\IEEEauthorrefmark{4}, Alexey Vinel\IEEEauthorrefmark{2}}
    \thanks{\IEEEauthorrefmark{1}XITASO GmbH, Augsburg, Germany. Email: \{lei.wan, hannan.keen\}@xitaso.com}
    \thanks{\IEEEauthorrefmark{2}Karlsruhe Institute of Technology, Karlsruhe, Germany. Email: alexey.vinel@kit.edu}
    \thanks{\IEEEauthorrefmark{3}LiangDao GmbH, Munich, Germany. Email: \{prabesh.gupta, andreas.eich\}@liangdao.de}
    \thanks{\IEEEauthorrefmark{4}Fraunhofer IVI, Dresden/Ingolstadt, Germany. Email: \{michael.kloeppel, marcel.kettelgerdes, maximilian.bialdyga\}@ivi.fraunhofer.de}
}

\maketitle

\begin{abstract}
Reliable perception remains a key challenge for \acf{CAVs} in complex real-world environments, where varying lighting conditions and adverse weather degrade sensing performance. While existing multi-sensor solutions improve local robustness, they remain constrained by limited sensing range, line-of-sight occlusions, and sensor failures on individual vehicles. This paper introduces VALISENS, a validated cooperative perception system that extends multi-sensor fusion beyond a single vehicle through \acf{V2X}-enabled collaboration between \acf{CAVs} and intelligent infrastructure. VALISENS integrates onboard and roadside LiDARs, radars, RGB cameras, and thermal cameras within a unified multi-agent perception framework. Thermal cameras enhances the detection of \acf{VRUs} under challenging lighting conditions, while roadside sensors reduce occlusions and expand the effective perception range. In addition, an integrated sensor monitoring module continuously assesses sensor health and detects anomalies before system degradation occurs. The proposed system is implemented and evaluated in a dedicated real-world testbed. Experimental results show that VALISENS improves pedestrian situational awareness by up to 18\% compared with vehicle-only sensing, while the sensor monitoring module achieves over 97\% accuracy, demonstrating its effectiveness and its potential to support future \ac{C-ITS} applications.
\end{abstract}


\section{INTRODUCTION} \label{introduction}

\begin{figure*}[htbp]
    \centering
    \begin{subfigure}[t]{0.26\textwidth}
        \centering
        \includegraphics[width=\linewidth]{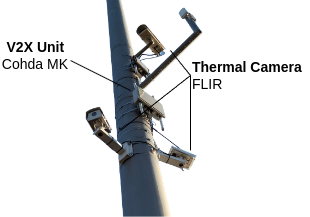}
        \caption{Camera node}
        \label{fig:camera-node}
    \end{subfigure}\hfill
    \begin{subfigure}[t]{0.26\textwidth}
        \centering
        \includegraphics[width=\linewidth]{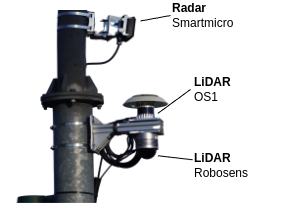}
        \caption{LiDAR and radar nodes}
        \label{fig:lidar-radar-node}
    \end{subfigure}\hfill
    \begin{subfigure}[t]{0.29\textwidth}
        \centering
        \includegraphics[width=\linewidth]{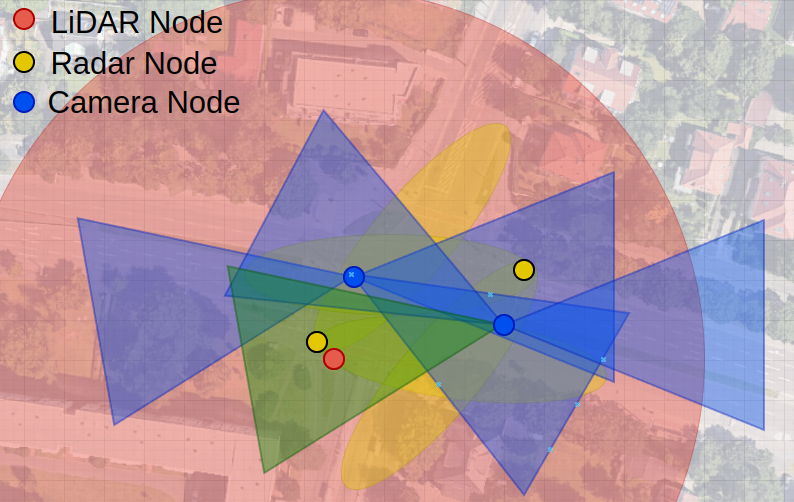}
        \caption{Roadside sensor layout}
        \label{fig:roadside-layout}
    \end{subfigure}

    \vspace{0.4cm}
    \hspace*{\fill}
    \begin{subfigure}[t]{0.29\textwidth}
        \centering
        \includegraphics[width=\linewidth]{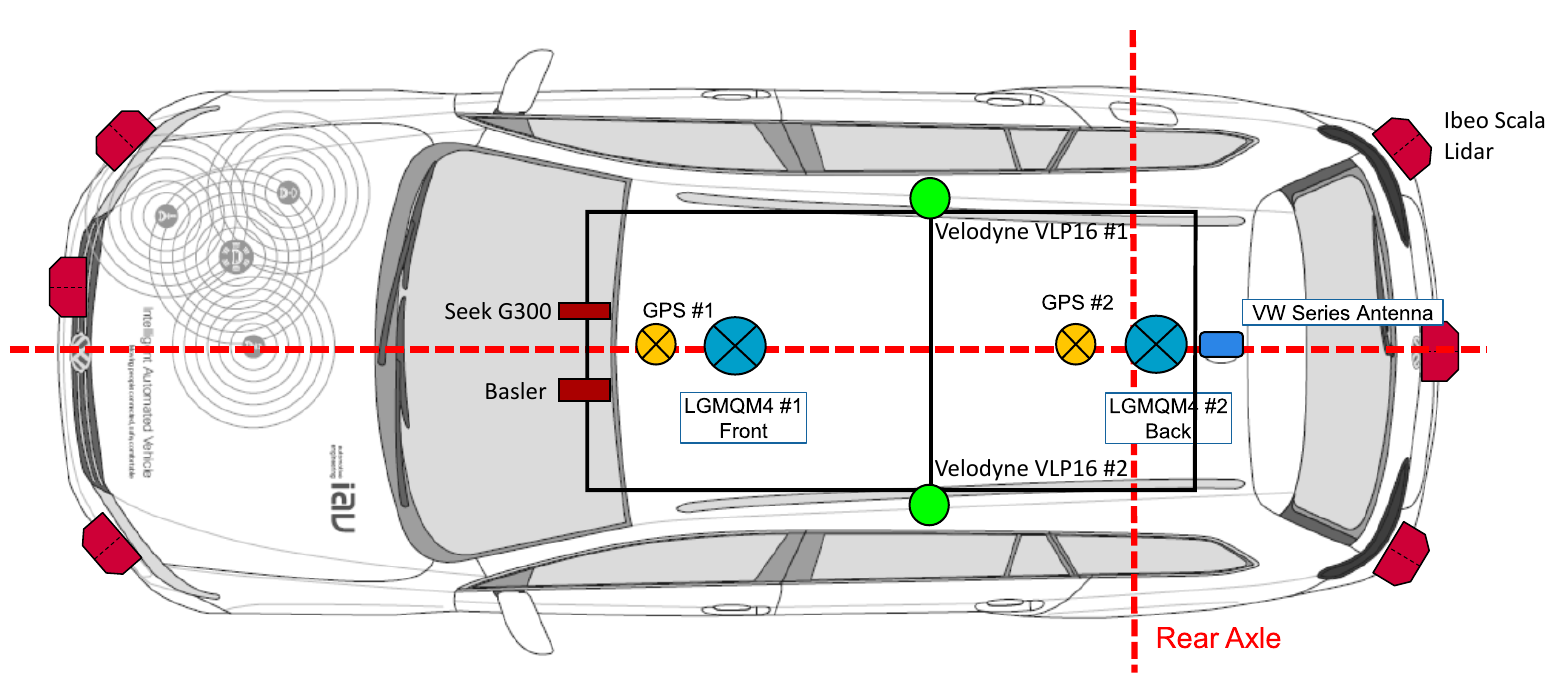}
        \caption{Sensor positions on test vehicle One}
        \label{fig:vehicle-one}
    \end{subfigure}\hfill
    \begin{subfigure}[t]{0.29\textwidth}
        \centering
        \includegraphics[width=\linewidth]{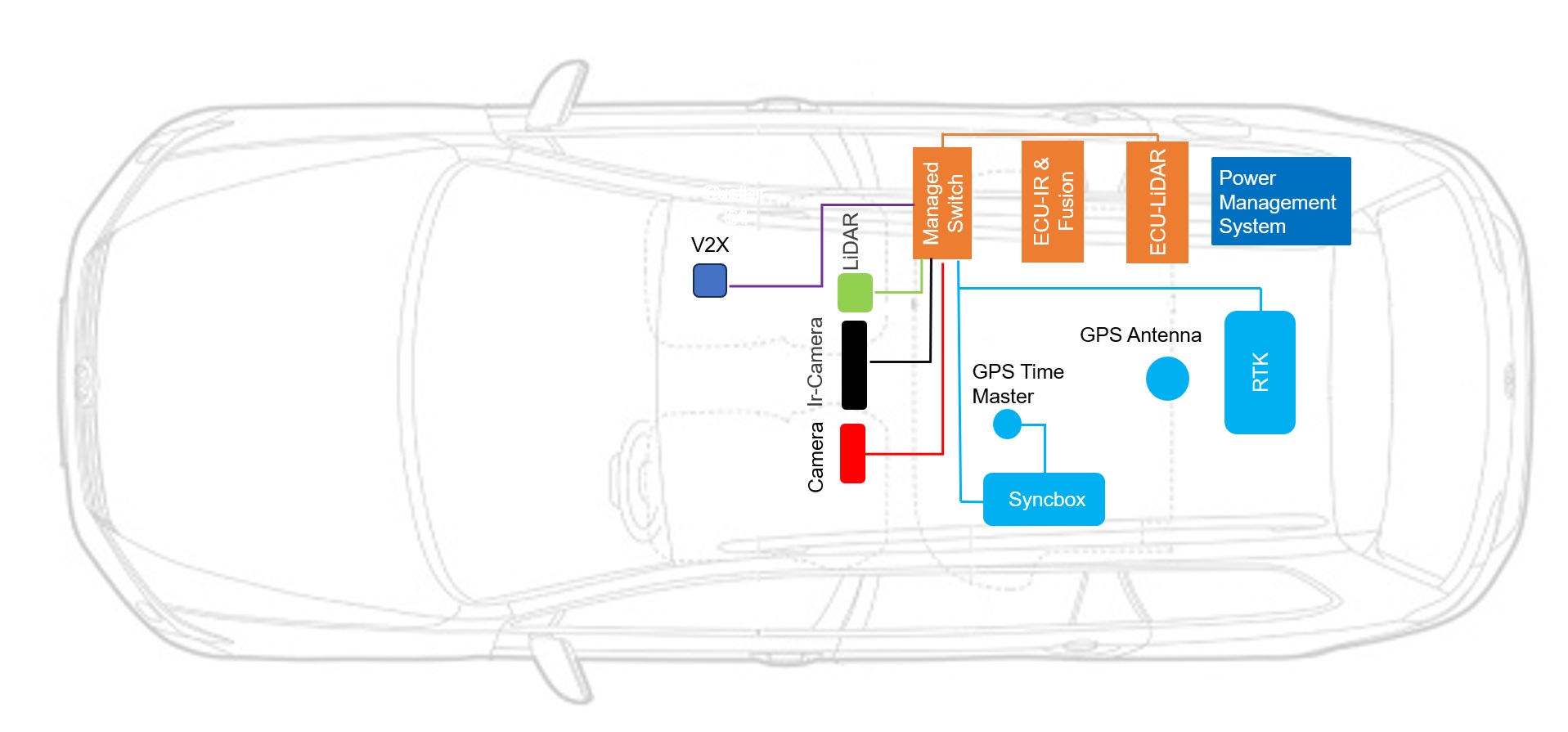}
        \caption{Sensor positions on test vehicle Two}
        \label{fig:vehicle-two}
    \end{subfigure}
    \hspace*{\fill}
    \caption{Roadside and onboard sensor configurations of VALISENS. The first row shows the roadside sensing infrastructure, including camera, LiDAR, and radar nodes, as well as their deployment layout at the intersection. The second row presents the sensor configurations of the two test vehicles used in the experiments.}
    \label{fig:valisens-sensor-layout}
\end{figure*}

The transportation system is undergoing a significant revolution, from isolated, manually operated vehicles and infrastructure toward a highly digitalized, intelligent, and connected ecosystem. At the core of this evolution are \acf{C-ITS}, which integrate \acf{CAV}s with intelligent infrastructure. These components are increasingly equipped with advanced sensors, high-performance computing units, and \acf{V2X} communication units, enabling them to perceive their surroundings and exchange data in real time. This shared situational awareness improves both autonomous navigation and infrastructure-supported transportation management.

To achieve robust perception, \ac{CAV}s typically rely on a combination of heterogeneous sensors, including LiDAR, radar, and RGB cameras \cite{10104104}. LiDAR provides accurate 3D spatial measurements, radar offers robustness in adverse weather conditions, and RGB cameras contribute rich semantic information. Through multi-sensor fusion, vehicles can maintain reliable perception across varying environment, such as different lighting or weather conditions. However, the perception capabilities of an individual vehicle remain constrained in complex traffic environments, particularly due to limited sensing range and occlusions caused by other objects \cite{wan2025components}. Recent advancements in \ac{V2X} communication have made it possible to extend perception beyond the individual vehicle by enabling the sharing of sensing data among vehicles and infrastructure \cite{wan2025systematic}. This \ac{CP} paradigm significantly enhances the detection range and mitigates the impact of visual occlusions, contributing to improved safety and traffic efficiency. As a result, a dynamic, decentralized sensor network can be established, comprising multiple sensor-equipped entities, that supports a comprehensive and continuously updated understanding of the surrounding environment.

This paper presents VALISENS, an integrated and validated sensor system designed to support cooperative automated driving. As shown in the Figure \ref{fig:valisens-sensor-layout}, the system combines on-board and roadside sensing units, including LiDARs, radars, RGB cameras, and thermal cameras, to enhance perception capabilities in complex urban environments. The main contributions of this paper are as follows:

\begin{itemize}
\item We design, implement, and deploy a cooperative perception system in a real-world testbed involving two \ac{CAV}s and two infrastructure units. The proposed system improves the situational awareness of \ac{VRU}s by 18\% compared with vehicle-only sensing.
\item We evaluate the perception performance using both public datasets and real-world experiments. Quantitative and qualitative results demonstrate the benefits of incorporating thermal cameras and roadside sensors into cooperative perception.
\item We validate the performance of the integrated sensor monitoring and \ac{V2X} communication modules in real-world deployment. The sensor monitoring module continuously assesses sensor health and supports early anomaly detection with accuracy over 97\%, enhancing system reliability.
\end{itemize}

\subsection{Related Work}

\subsubsection{Perception}

Perception for autonomous vehicles has been extensively studied in recent years, evolving from single-sensor approaches to multi-sensor fusion systems, and from isolated in-vehicle or roadside sensors to dynamic, multi-agent sensor networks. In-vehicle sensor systems have been the foundation of many large-scale vehicle-perspective datasets \cite{geiger2013vision,caesar2020nuscenes,sun2020scalability,9636848}, supporting a wide range of perception tasks including object detection, tracking, semantic segmentation, and motion prediction. Recently, there has been a growing interest in roadside sensor systems, as reflected in the increasing number of roadside perception datasets \cite{ye2022rope3d,10422289,mirlach2025r}. The integration of sensor data from both vehicles and infrastructure has led to the development of \ac{CP} datasets such as DAIR-V2X \cite{dair-v2x}, V2X-Seq \cite{v2x-seq}, V2V4Real \cite{xu2023v2v4real}, and V2X-Real \cite{xiang2024v2x}. Collectively, these datasets underscore the growing focus on multi-agent sensor networks that incorporate diverse modalities across vehicles and infrastructure.

In such networks, data fusion is a key component for generating a unified and consistent understanding of the environment. Fusion strategies are commonly categorized into early fusion \cite{JointPerceptionScheme-2022-ahmeda}, intermediate fusion \cite{SlimFCPLightweightFeatureBasedCooperative-2022-guoa}, and late fusion \cite{10161460}. Among these strategies, late fusion, which exchanges and aggregates object-level information, is considered the most practical solution for scalable real-world deployment. It is model- and modality-agnostic and significantly reduces bandwidth requirements compared to early fusion, making it better aligned with the realistic constraints of \ac{V2X} communication. The system proposed in this paper adopts a late fusion approach to enable modular perception across heterogeneous sensors and ensure compatibility with communication standards defined by \ac{ETSI}.

\subsubsection{Sensor Condition Monitoring}
The performance of a multi-modal perception system is directly dependent on error-free operation and data integrity of the underlying sensor hardware. As compiled by Goelles et al.~\cite{Goelles.2020} and Secci et al.~\cite{Secci.2020}, especially in the case of optical sensors such as cameras and LiDAR, a multitude of failure mechanisms can, however, lead to severely degraded performance in terms of output data quality or even total system failure.    
Thus, common error sources include well-studied external effects, such as sensor blockage due to leaves and dirt~\cite{Trierweiler.102019}, but also adverse weather conditions such as rain, fog or sun glare~\cite{Linnhoff.2022,Brophy.2023}. 
In addition to external effects, internal failure or component degradation might occur over time, especially considering the demanding thermo-mechanical operating conditions in the automotive domain. Internal effects comprise electrical faults, including dead pixels, as well as component failures, such as malfunctioning processing units and LiDAR laser diodes~\cite{Secci.2020,Goelles.2020}. Furthermore, it could be shown that the sensor's optical receiver front end can be prone to degradation, expressed, for example, in intrinsic decalibration~\cite{Li.2022,Kettelgerdes.2021b} or loss of sharpness~\cite{Pandey.2023}.

Considering the multitude of possible failure sources and their high criticality due to the potentially hazardous consequences in safety-critical automotive applications, condition monitoring of these sensors plays a crucial role in robust environmental perception. However, while there exist several works on the monitoring of external effects, such as weather conditions~\cite{Dhananjaya.2021,Kettelgerdes.2024b}, only a few focus on internal effects. Plus, existing works propose rather abstract monitoring concepts and frameworks than specific methods, which could be due to the lack of publicly available data in this domain, as noted by Goelles et al.~\cite{Goelles.2020}. Following that, extensive reliability testing of sensors constitutes a crucial part of the overall condition monitoring methodology~\cite{Kettelgerdes.2024}.  
\subsubsection{Communication}

Communication plays a central role in \ac{C-ITS}, enabling real-time cooperation between vehicles, infrastructure, and other road users. \ac{DSRC} and cellular networks are two primary technologies used for message exchange \cite{jin2022dsrc}. \ac{DSRC} allows direct communication between devices with low latency, making it suitable for safety-critical applications \cite{mannoni2019comparison}. However, its limited bandwidth restricts the range of supported services. To address these limitations, \ac{C-V2X} has emerged as a complementary technology, offering greater bandwidth and supporting data-intensive applications such as \ac{CP}. To enhance interoperability and deployment flexibility, dual-mode V2X modules supporting both \ac{DSRC} and \ac{C-V2X} are now commercially available \cite{8926330}. The system proposed in this paper uses such a dual compatible V2X stack to facilitate testing of both communication protocols and to ensure interoperability between heterogeneous vehicles from different manufacturers.

\section{SYSTEM ARCHITECTURE} \label{system}

This section provides an overview of the proposed multi-agent sensor system and the core functionalities built upon it. Section~\ref{sys_overview} describes the sensor-equipped infrastructure and vehicle platforms. Sections~\ref{env_perception}, and \ref{sensor_monitoring} present the key functional modules of the system, including environment perception and sensor monitoring.

\subsection{Overview} \label{sys_overview}
The overall system comprises both infrastructure and vehicles, each equipped with multiple sensors, computing units, and \ac{V2X} communication stack, as detailed in the following sections. Intra-entity communication is handled using ROS2\cite{ROS2}, while inter-entity communication, specifically between infrastructure and vehicles, is facilitated through \ac{V2X}.

\begin{figure}[!t]
    \centering
    \includegraphics[width=0.53\linewidth]{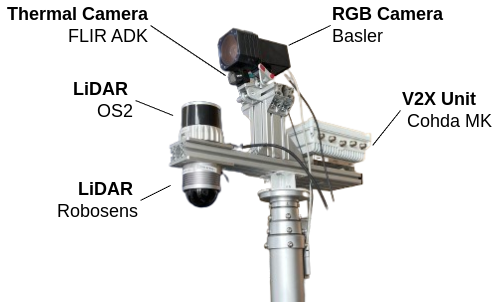}
    \caption{Sensor configurations of the mobile roadside sensor platform}
    \label{fig:mobile-platform}
\end{figure}

\subsubsection{Infrastructure systems}
Two test beds are utilized to support the sensor system. The primary test bed is located at a public intersection in Dresden, Germany (Zellescher Weg/Paradiesstraße). As Figure \ref{fig:valisens-sensor-layout} shows, this site is equipped with six cameras (five thermal cameras and one dual-mode RGB/thermal camera), one LiDAR system (Ouster OS1 with 64 layers in below horizon configuration, complemented, in the blind-zone, by a Robosense BPearl), and one radar system featuring 4×4 MIMO capability. All roadside sensors are synchronized using an \ac{NTP} server provided by the \ac{RSU} and are calibrated to a shared intersection base link. For data processing, three NVIDIA Jetson AGX Orin devices and one Jetson AGX Xavier are deployed and connected to all sensor components via Ethernet. Communication with vehicles and back-end systems is facilitated by a Cohda MK6 \ac{RSU} for V2X messaging and a Milesight UR75 5G router. A detailed description of this test bed can be found in~\cite{kloppel2021fraunhofer}. 

The second test bed is located in Ingolstadt and consists of 22 sensor masts. The western part of the test bed is characterized by a heterogeneous sensor setup, where only three of the eleven masts are equipped with V2X communication.  In contrast, the eastern part is characterized by a homogeneous sensor and communication setup, consisting of LiDAR systems (the same setup as in the Dresden test bed), RGB cameras, and Cohda MK6. A detailed description of the test bed, including V2X measurements, can be found in \cite{song2024first}. 

To complement the fixed infrastructure, a mobile sensor platform, referred to as the Rover, is also employed. The Rover, shown in Figure \ref{fig:mobile-platform}, is equipped with one RGB camera, one thermal camera, and the same LiDAR configuration as used in the fixed test beds. It is also equipped with a Cohda MK6 \ac{RSU} for V2X communication, a Jetson AGX Orin and an x86\_64-based computing unit with an NVIDIA RTX 4060 GPU for real-time data processing.

\subsubsection{Vehicle systems}
In parallel to the infrastructure setup, two sensor-equipped vehicles are used for testing and evaluation. As Figure \ref{fig:valisens-sensor-layout} shows, both vehicles feature a dual-camera setup consisting of a Seek infrared camera and a Basler RGB camera, as well as a Cohda MK6 \ac{OBU} for V2X communication. Vehicle One is outfitted with two Velodyne VLP-16 LiDARs mounted on the roof and six IBEO Scala LiDARs positioned on the front and rear bumpers. It also includes a NovAtel PowerPak7-E1 \ac{GNSS} with integrated IMU for high precision localization. The onboard computing platform comprises three x86\_64-based units, one of which is equipped with an NVIDIA RTX 2000 GPU to enable real-time AI processing. All onboard sensors are synchronized using an NTP server and calibrated relative to the Velodyne LiDAR. A prior version of this vehicle configuration is described in~\cite{vehits21}. 

Vehicle Two is equipped with an InnovizOne LiDAR mounted on the roof and a DTC xNAV650 (GPS/RTK) inertial navigation system. For onboard processing, it features two Jetson AGX Orin devices, one dedicated to image processing and the other to LiDAR data processing. The sensors in Vehicle Two are synchronized via a \ac{PTP} server and calibrated with respect to the LiDAR system. 

\subsection{Environmental perception concept} \label{env_perception}
\begin{figure}[t]
    \centering
    \includegraphics[width=0.95\linewidth]{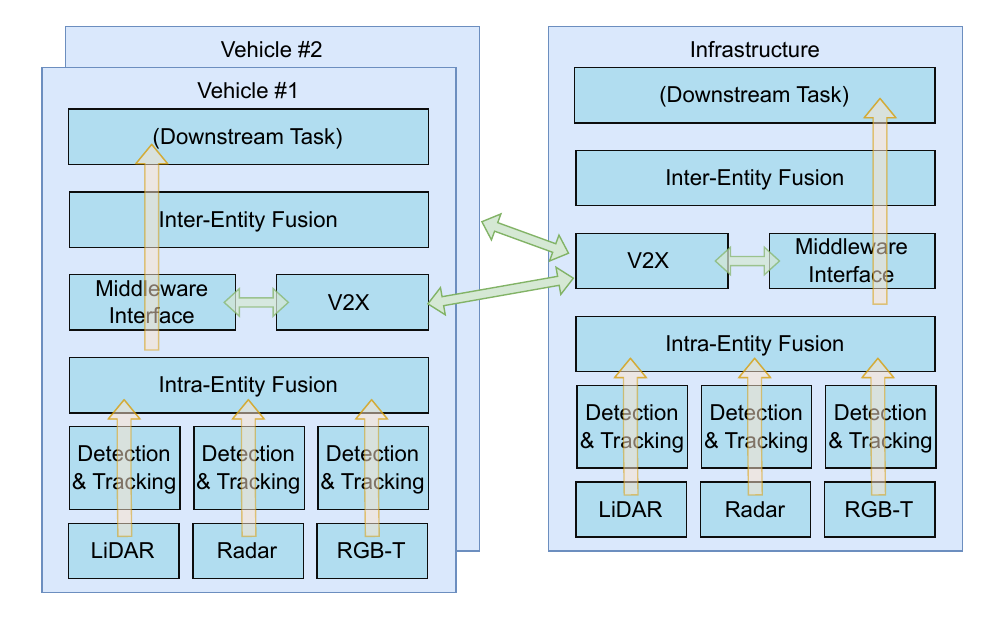}
    \caption{Software architecture overview of environmental perception.}
    \label{fig:software_architecture}
\end{figure}

The concept of environmental perception is modularized by the sensor modality, as illustrated in Figure~\ref{fig:software_architecture}. Each modality independently delivers a list of detected and tracked objects to downstream modules, enabling the seamless integration of heterogeneous sensors. The perception outputs are first aggregated by the intra-entity fusion node, which combines data from multiple sensors within the same platform. Subsequently, the inter-entity fusion node integrates information across different agents (e.g., vehicles and infrastructure). The fused object-level data is then passed to the downstream module to support cooperative automated driving.

\subsubsection{LiDAR}

In the proposed system, LiDAR data is processed using a LiangDao's in-house developed AI-based perception algorithm to enable both short- and long-range detection and tracking of relevant traffic participants. The output includes 3D bounding boxes with associated semantic labels, tracking ID, 3D position, 3D dimensions, velocity, and heading angle. LiDARs are deployed across both infrastructure and vehicle platforms in various configurations. In intra-entity setups involving multiple LiDAR units, point clouds are fused prior to perception module using an early fusion approach, thereby enabling a more complete and consistent view of the environment.
  
\subsubsection{Radar}

Multiple radar units are installed on infrastructure, each oriented toward different road directions to ensure comprehensive coverage of the intersection area. The outputs of the single radar sensors are fused by an in-house algorithm by smartmicro, which returns object lists represented in a \ac{BEV}, excluding height information (z-dimension).  To enable fusion with LiDAR data, radar object lists undergo post-processing by assigning a default height value, thereby reconstructing the data into full 3D coordinates.

\subsubsection{RGB-T}

In the proposed architecture, RGB and thermal data are processed independently and fused using a late fusion strategy; even if one camera failed, the RGB-T module still detect and track the objects. Here, we utilize the YOLO framework \cite{khanam2024yolov11} as the detector for each camera, producing a 2D object bounding box in the image plane. To aggregate results from multiple cameras, the bounding boxes are first projected to the core camera image plane and then fused by Non-Maximum Suppression (NMS) \cite{neubeck2006efficient} operation. The resulting fused bounding boxes are then passed to a 2D multi-object tracking node, which employs the ByteTrack algorithm \cite{zhang2022bytetrack} to track objects across frames in real time.

\subsubsection{Intra-Entity Fusion} \label{sensor_fusion}
To produce a unified and consistent perception output, all modality-specific results are forwarded to the intra-entity fusion node. The proposed system employs a late fusion approach that integrates object-level information from multiple sensors.

\begin{figure}[t]
    \centering
    \begin{subfigure}[b]{0.48\linewidth}
        \centering
        \includegraphics[width=\linewidth]{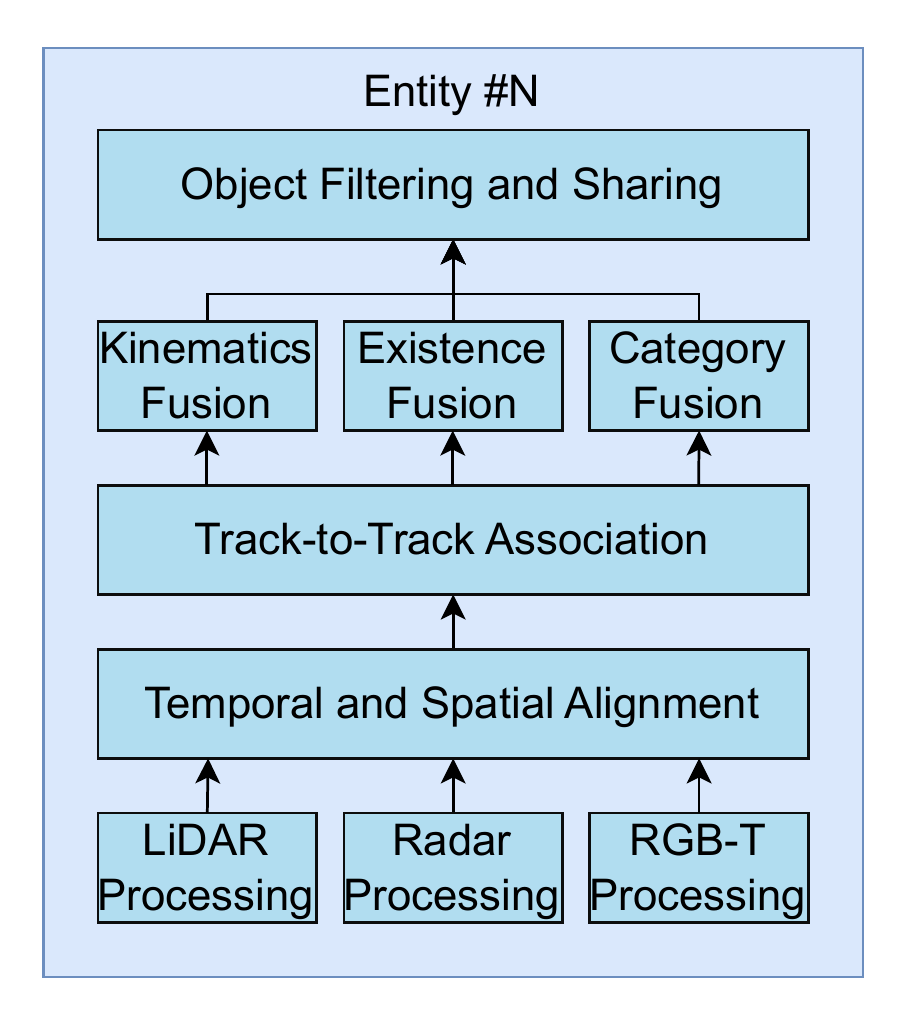}
        \caption{Intra-entity fusion}
        \label{fig:intra_fusion}
    \end{subfigure}
    \begin{subfigure}[b]{0.43\linewidth}
        \centering
        \includegraphics[width=\linewidth]{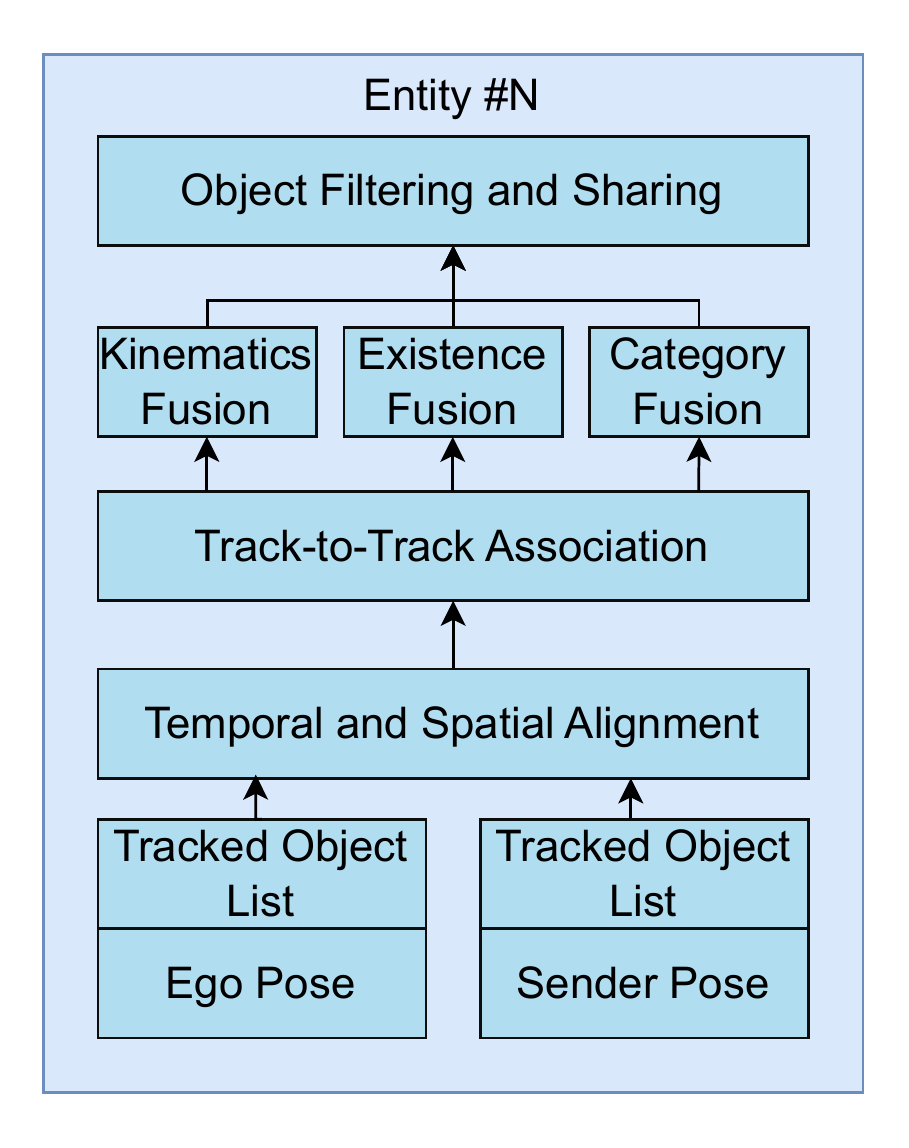}
        \caption{Inter-entity fusion}
        \label{fig:inter_fusion}
    \end{subfigure}
    \caption{Schematic diagrams of intra-entity and inter-entity fusion approaches}
    \label{fig:fusion}
\end{figure}

As shown in Fig. \ref{fig:fusion}, the fusion module utilizes a track-to-track fusion algorithm \cite{6222334}, which consists of three main stages: alignment, association, and fusion. In the \textit{alignment stage}, the object data is spatially transformed from the local coordinate system of each sensor into the vehicle’s base-link coordinate. Temporal alignment is achieved using a constant velocity model to project object states to a common timestamp. For LiDAR–camera fusion, 3D bounding boxes from LiDAR are projected onto the image plane to associate them with corresponding 2D detections from the camera. Once aligned, \textit{association stage} is performed using a combination of Euclidean distance and Intersection-over-Union (IoU) thresholds to match corresponding detections across modalities. In \textit{fusion stage}, associated objects are then fused using an adapted Kalman filter, where each observation is treated as a measurement to update the global object state and covariance. In addition to position and velocity, the fusion process also incorporates object classification and shape. Class labels from camera are given higher weight due to their superior semantic accuracy, while LiDAR-based estimates are prioritized for object shape due to their precise 3D measurement. Radar-based velocity estimates are given a higher weight because of their direct Doppler-based measurement mechanism, which provides accurate motion information. Before publishing the final fused output, object existence probabilities are recalculated, and only those exceeding a predefined confidence threshold are retained.

\begin{table}[t]
\centering
\small
\caption{RGB-T object detection results on mobile sensor platform, $\text{AR}^{100}_{\text{per}}$ denotes average recall for pedestrians considering up to 100 detections.}
\setlength{\tabcolsep}{5pt}
\begin{tabular}{llcccc}
\toprule
\multirow{2}{*}{Model} & \multirow{2}{*}{Modality}  & \multicolumn{2}{c}{Day} & \multicolumn{2}{c}{Night} \\
\cmidrule(lr){3-4} \cmidrule(lr){5-6}
&  & $\text{AP}_{50}$ & $\text{AR}^{100}_{\text{per}}$ & $\text{AP}_{50}$ & $\text{AR}^{100}_{\text{per}}$ \\
\midrule
\multirow{3}{*}{YOLOv8 \cite{Ultralytics}} 
    & RGB    & \textbf{63.02} & 46.39 & 49.85 & 37.76 \\
    & Thermal     & 50.11 & 33.39 & 49.04 & 31.83 \\
    & RGB+T & 59.93 & \textbf{50.42} & \textbf{53.26} & \textbf{45.34}\\
\bottomrule
\end{tabular}
\label{tab:rgbt_detection}
\end{table}

\begin{table*}[h]
\centering
\caption{Perception Performance on V2X-Seq-SPD.
$AR_{100}$ denotes average recall with up to 100 detections per frame.}
\label{tab:multiclass_all}
\setlength{\tabcolsep}{5.0pt}
\begin{tabular}{lcccccccccccc}
\toprule
 & \multicolumn{4}{c}{Car} & \multicolumn{4}{c}{Cyclist} & \multicolumn{4}{c}{Pedestrian} \\
\cmidrule(lr){2-5} \cmidrule(lr){6-9} \cmidrule(lr){10-13}
Method 
& $AP_{50}$ & $AR_{100}$ & AMOTA & AMOTP$\downarrow$
& $AP_{50}$ & $AR_{100}$ & AMOTA & AMOTP$\downarrow$
& $AP_{50}$ & $AR_{100}$ & AMOTA & AMOTP$\downarrow$ \\
\midrule
Vehicle
& 68.43 & \textbf{69.82} & 0.503 & \textbf{0.957}
& 50.87 & \textbf{66.19} & 0.134 & 0.851
& 23.60 & 65.22 & 0.035 & 0.691 \\
Intra-fusion
& 68.53 & 68.59 & 0.625 & \textbf{0.957}
& 48.95 & 59.26 & \textbf{0.285} & \textbf{0.848}
& 24.83 & 59.00 & 0.020 & \textbf{0.637} \\
Inter-fusion
& \textbf{68.66} & 68.37 & \textbf{0.627} & 0.960
& \textbf{50.88} & 63.83 & 0.186 & 0.852
& \textbf{30.86} & \textbf{66.53} & \textbf{0.091} & 0.692 \\
\bottomrule
\end{tabular}
\label{tab:dair_v2x_result}
\end{table*} 

\subsubsection{Communication and Inter-Entity Fusion} 

The proposed system adopts a late fusion strategy for inter-entity fusion, in alignment with the communication standards defined by \ac{ETSI}. Specifically, the system utilizes the \ac{CPM} \cite{CPM23} message to transmit perception data, which include the pose of the ego vehicle along with a list of tracked objects. In the implemented architecture, the output of the intra-entity fusion module is used to generate CPM messages, which are then broadcast via the \ac{V2X} communication stack. Other entities within the communication range receive and decode the CPM, extracting both the sender’s pose and its associated perception results. To fuse data from external agents, the same fusion strategy described in Section~\ref{sensor_fusion} is applied. Received objects are first transformed from the sender's coordinate frame into the ego vehicle’s frame. They are then spatially and temporally aligned, associated with locally perceived objects, and fused using the track-to-track fusion pipeline, as shown in Figure \ref{fig:fusion}. This process enables robust and seamless cooperation among heterogeneous agents, resulting in a continuous and unified understanding of the surrounding environment.

\subsection{Sensor Condition Monitoring Concept} \label{sensor_monitoring}

In order to ensure reliable perception performance and resilience with respect to failure sources introduced in Sec.~\ref{introduction}, a sensor condition monitoring module is developed and evaluated for automotive grade camera sensors in the test bed Ingolstadt. For that, extensive reliability testing is conducted in order to identify potential internal failures in addition to well-studied external effects. Initial thermal cycling tests, conducted under the LV124 automotive test standard for accelerated aging~\cite{Volkswagen.2009}, did not show a significant gradual degradation of optical performance in the form of loss of sharpness or decalibration. Hence, the condition monitoring concept focuses on anomaly detection of permanent and non-permanent external and internal faults, such as adverse weather conditions, blockage by dirt, ice, or fogging, and permanent imager faults.

\section{EXPERIMENTS AND RESULTS} \label{results}
This section evaluates the proposed VALISENS system from both single-entity and multi-entity perspectives. We first quantify the benefits of intra-entity and inter-entity fusion on public benchmarks, and then validate these findings through real-world cooperative driving experiments.

\begin{table*}[h]
\centering
\caption{Perception Performance of the VALISENS System in Real-World Experiments.
$AR_{100}$ denotes average recall with up to 100 detections per frame.}
\label{tab:valisens_all}
\setlength{\tabcolsep}{5.0pt}
\begin{tabular}{lcccccccccccc}
\toprule
 & \multicolumn{4}{c}{Car} & \multicolumn{4}{c}{Cyclist} & \multicolumn{4}{c}{Pedestrian} \\
\cmidrule(lr){2-5} \cmidrule(lr){6-9} \cmidrule(lr){10-13}
Method 
& $AP_{50}$ & $AR_{100}$ & AMOTA & AMOTP$\downarrow$
& $AP_{50}$ & $AR_{100}$ & AMOTA & AMOTP$\downarrow$
& $AP_{50}$ & $AR_{100}$ & AMOTA & AMOTP$\downarrow$ \\
\midrule
Vehicle
& \textbf{73.94} & \textbf{89.30} & 0.342 & \textbf{0.834}
& 12.50 & 20.31 & \textbf{0.258} & \textbf{0.457}
& 55.57 & \textbf{98.13} & \textbf{0.487} & 0.707 \\
Intra-fusion
& 73.63 & 85.02 & 0.352 & 0.837
& 12.50 & 20.31 & 0.217 & \textbf{0.457}
& 65.25 & 97.61 & 0.323 & 0.704 \\
Inter-fusion
& 73.87 & 86.47 & \textbf{0.353} & 0.896
& 12.50 & \textbf{23.52} & 0.220 & 0.459
& \textbf{65.75} & 97.90 & 0.363 & \textbf{0.688} \\
\bottomrule
\end{tabular}
\end{table*}

\begin{figure*}[h]
    \centering
        \includegraphics[width=\linewidth]{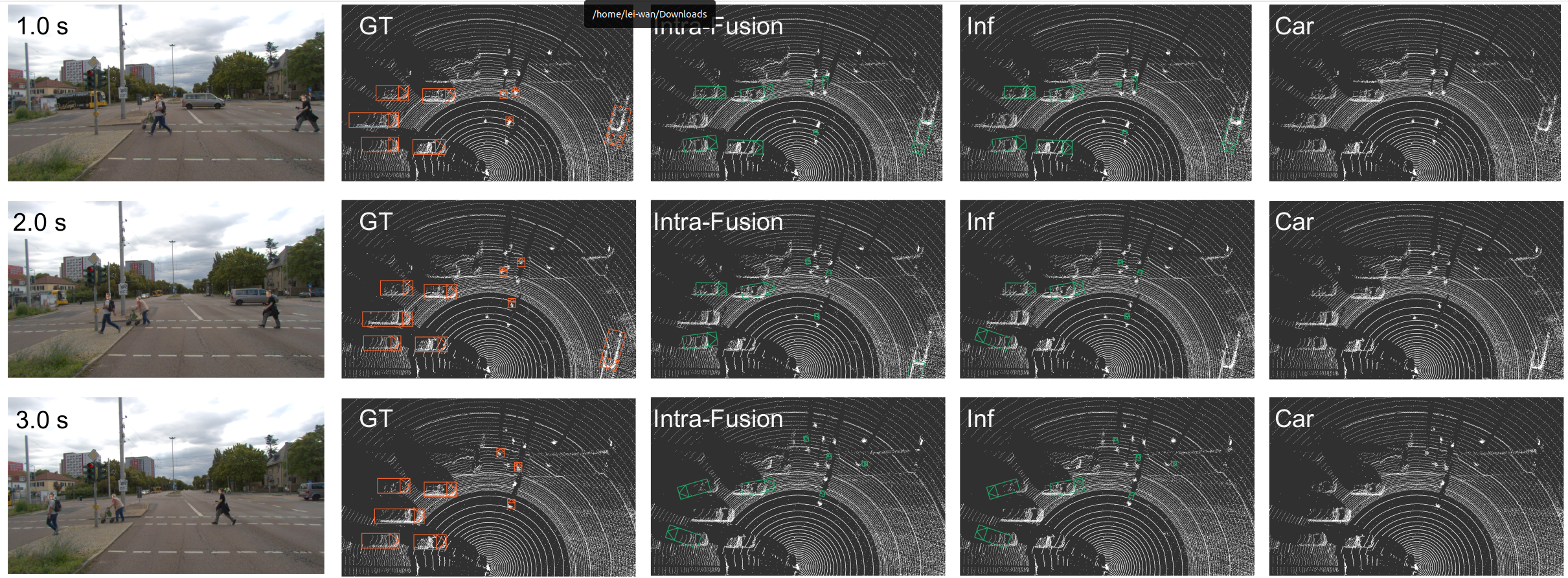}
        \caption{Qualitative comparison of perception results under different fusion modes in the testbed. Each row shows a snapshot of an unprotected left-turn scenario at different time instances. RGB images are captured from a vehicle not participating in intra-entity fusion, providing a clearer view of the pedestrian. Intra-entity fusion demonstrates more reliable detection and tracking of \ac{VRU} compared with single-source fusion modes.}
        \label{fig:valisens_result}
\end{figure*}

\subsubsection{Results on Public dataset}

Our public-dataset evaluation focuses on two questions:
(i) whether multi-modal fusion improves perception on a single vehicle (intra-entity fusion), and
(ii) whether cooperative perception further enhances detection and tracking performance (inter-entity fusion). To answer these questions, we use the R-LiViT \cite{mirlach2025r} and V2X-Seq \cite{v2x-seq} datasets. 

\textbf{Intra-entity RGB–Thermal fusion.}
R-LiViT contains 10,000 LiDAR frames and 2,400 paired RGB–thermal images collected using the mobile sensing platform shown in Fig.~\ref{fig:mobile-platform}. Table~\ref{tab:rgbt_detection} reports detection results under different illumination conditions. Compared with RGB-only perception \cite{FasterRCNN,Ultralytics,RT-DETR}, RGB–thermal fusion improves nighttime $AP_{50}$ from 49.85 to 53.26. Pedestrian recall is consistently higher in both daytime and nighttime scenarios. These results indicate that thermal sensing effectively complements RGB cameras, especially for \ac{VRU} under low-visibility conditions.

\textbf{Intra-entity LiDAR–Camera Fusion.}
Due to imperfect temporal alignment between LiDAR and camera data in the R-LiViT dataset, we additionally use the V2X-Seq-SPD dataset to evaluate intra-entity fusion. LiDAR-based perception is implemented using PointPillars \cite{lang2019pointpillars} with AB3DMOT \cite{Weng2020_AB3DMOT}, while camera-based perception employs YOLOv12 \cite{Ultralytics} with ByteTrack \cite{zhang2022bytetrack}. As shown in Table~\ref{tab:dair_v2x_result}, fusing LiDAR and camera data significantly improves \ac{VRU} perception compared with LiDAR-only baselines. For example, cyclist AMOTA increases from 0.134 to 0.285, and pedestrian $AP_{50}$ improves from 23.60 to 24.83. These gains demonstrate that camera information is particularly beneficial for detecting small and distant objects.

\textbf{Inter-entity Fusion.}
Building on intra-entity fusion, we introduce vehicle–infrastructure cooperation. As shown in Table~\ref{tab:dair_v2x_result}, compared with the vehicle-only pipeline, inter-entity fusion increases pedestrian $AP_{50}$ by approximately 30\%, reaching 30.86. These results show that inter-entity fusion substantially extends sensing coverage and reduces missed detections, highlighting its importance for  \ac{VRU} protection as well. Although tracking performance improves from an AMOTA of 0.03 to 0.09 after intra-entity fusion, pedestrian tracking performance remains limited and needs to be addressed.

\subsubsection{Real-world experiments}
To verify the generalizability of the above findings, we conduct real-world cooperative driving experiments under multiple scenarios, including car following, straight driving, and left and right turns. We manually annotate more than 2,400 frames across four sequences and evaluate detection and tracking performance for cars, cyclists, and pedestrians.

Consistent with the public-dataset results, both intra-entity and inter-entity fusion improve VRU perception in real traffic. Inter-entity fusion achieves the best performance, with pedestrian $AP_{50}$ reaching 65.75 and an average recall of 97.60. The high recall indicates that most pedestrians are detected with very few missed instances, demonstrating the practical safety benefits of the proposed cooperative perception system. Representative qualitative results are shown in Fig.~\ref{fig:valisens_result}, where cooperative fusion enables stable detection and tracking of VRUs that are missed by single-vehicle perception pipelines.

Beyond perception accuracy, communication performance is a key factor affecting the effectiveness of inter-entity fusion. We therefore conduct multi-round measurements in the real-world testbed to quantify communication quality and capacity. Measurements were collected over a three-hour period around noon, with statistics computed from more than 2,700 exchanged messages, including \acp{CAM} and \acp{CPM}. Table \ref{tab:communication} summarizes key communication performance indicators. The high download values for the vehicle were achieved by using Multi-WAN, i.e. the router is connected to and uses two mobile communication providers in parallel. The V2X communication range was evaluated from \acp{CAM} \cite{CAM2019}, achieving a maximum range exceeding 1~km. Since these messages are not forwarded, the embedded location information can be used to estimate the effective communication range. The reported values correspond to line-of-sight conditions and therefore represent a best-case scenario.

Furthermore, even when 81 detected objects are reported at the maximum update frequency of \SI{10}{\hertz}, only 33~kB are transmitted per second for the exchange of \acp{CPM}, while the average data rate is significantly lower than this value. Next, the one way delay for \ac{V2X} messages was evaluated by measuring the send and receive times, with maximum value of 5.0 ms. For this, both the sender and the receiver were calibrated using \ac{GNSS} time with \ac{PPS}.

\begin{table}[h]
    \centering
    \caption{Communication Performance Indicators.}
    \begin{tabular}{lc}
        \hline
        Modality & Measurements\\
        \hline
         5G Dresden test bed & UL: $\SI{80}{Mbps}$, DL: $\SI{65}{Mbps}$ \\
         5G Vehicle One & UL: $\SI{65}{Mbps}$, DL: $\SI{0.6}{Gbps}$\\
         V2X Dresden test bed & Max. range: $\geq \SI{1}{\kilo\metre}$ \\
         V2X Dresden test bed & Data rate: $\SI{33}{Kbps}$ max, $\SI{13}{Kbps}$ avg\\
         V2X Dresden test bed & One way delay: $\SI{5.0}{\milli\second}$ max, $\SI{3.0}{\milli\second}$ avg\\
         \hline
    \end{tabular}
    \label{tab:communication}
\end{table}

\subsection{Sensor condition monitoring}
Reliable sensor system requires that sensor faults be detected and handled before they propagate into higher-level fusion. We therefore develop and evaluate a camera sensor condition monitoring module designed for automotive-grade cameras operating under both external and internal failure sources.

Initial accelerated aging experiments following LV124 thermal cycling did not indicate significant gradual optical degradation, such as loss of sharpness or decalibration. Consequently, the proposed monitoring focuses on detecting permanent and non-permanent anomalies that can immediately affect perception quality, including adverse weather, lens blockage caused by dirt, ice, or fogging, and permanent imager defects. 

To train a robust anomaly detector, we combine reliability-test recordings with labelled public driving datasets \cite{wenzel20204seasons,burnett_ijrr23,sakaridis2018model,jimaging10110281} and complement them with physically induced faults to cover underrepresented classes. A classifier is trained directly on raw camera images to predict dedicated fault classes that can be communicated to higher-level fusion modules for mitigation and decision making. As shown in Table~\ref{tab:single_image_results}, we benchmark four candidate models. ConvNeXt~\cite{liu2022convnet} achieves the best performance with an F1-macro score of 0.981, followed closely by the Multi-Scale Vision Transformer (MViT)~\cite{fan2021multiscale} with 0.978. Both models demonstrate strong multi-label anomaly prediction capability; MViT was selected due to superior generalization on unseen datasets.

\begin{table}[t]
\centering
\caption{Performance comparison of anomaly classification models.Acc. denoates accuracy.}
\label{tab:single_image_results}
\renewcommand{\arraystretch}{1.15}
\setlength{\tabcolsep}{3.5pt}
\footnotesize
\begin{tabular}{lccccc}
\hline
Model &
F1$_{\mathrm{macro}}$ &
F1$_{\mathrm{micro}}$ &
Acc. &
Time (ms) &
Parameters (M) \\
\hline
SVM \cite{cortes1995support}     & 0.905 & 0.899 & 0.842 & 74.0  & N/A \\
ConvNeXt \cite{liu2022convnet} & \textbf{0.981} & \textbf{0.984} & \textbf{0.972} & \textbf{9.2} & 27.8 \\
MViT \cite{fan2021multiscale}     & 0.978 & 0.981 & 0.967 & 10.0  & \textbf{23.3} \\
Swin \cite{9710580}     & 0.918 & 0.933 & 0.908 & 10.7  & 27.5 \\
\hline
\end{tabular}
\end{table}

While this offline evaluation verifies anomaly detection accuracy, it does not directly quantify the impact of sensor anomalies on downstream perception. We therefore conduct on-road experiments on the public testbed under controlled lens conditions, including clear, broken, and partially covered lenses (leaf occlusion). As shown in Table \ref{tab:scene_results}, using a YOLO-based object detector, we observe a monotonic degradation in detection performance as sensor conditions deteriorate, with the mAP decreasing from 0.65 to 0.54 and further to 0.28, and the F1-score correspondingly declining from 0.74 to 0.59 and then to 0.34. These results establish a direct link between sensor anomalies and perception degradation, motivating actionable system responses such as driver warnings, cleaning requests, or fallback perception strategies.

\begin{table}[t]
\centering
\caption{Camera-based object detection performance under different sensor conditions in the testbed.}
\label{tab:scene_results}
\begin{tabular}{lccc}
\hline
Condition & Samples & F1-score & mAP \\
\hline
Clear            & 60                & 0.74     & 0.65 \\
Broken           & 50                & 0.59     & 0.54 \\
Covered          & 51                & 0.34     & 0.28 \\
\hline
\end{tabular}
\end{table}

\section{CONCLUSION} \label{conclusion}

This paper presents a multi-sensor system that integrates onboard and roadside sensing to support cooperative perception within \ac{C-ITS}. By combining data from multiple sources, \ac{CAV}s obtain a more complete and reliable understanding of their surroundings, even under adverse weather and lighting conditions. Roadside sensors reduce visual occlusions and extend the effective perception range beyond the capabilities of onboard sensors alone. Experimental results from the testbed demonstrate the potential of VALISENS for \ac{C-ITS} applications, including enhanced protection of \ac{VRU}s.

\bibliographystyle{IEEEtran}
\bibliography{IEEEabrv,bibliography}

\end{document}